\documentclass[runningheads]{llncs}
\usepackage[T1]{fontenc}
\usepackage{graphicx}
\usepackage[font=footnotesize,labelfont=footnotesize]{subfig}
\usepackage{amsmath}
\usepackage{booktabs}
\usepackage{caption}
\usepackage[margin=1.65in]{geometry}
\usepackage{wrapfig}
\usepackage{enumitem}
\usepackage{xcolor}
\usepackage{orcidlink}

\newcommand{\fref}[1]{Fig.~\ref{#1}}
\newcommand{\tref}[1]{Table~\ref{#1}}

\begin{document}
\title{Enabling Heterogeneous Adversarial Transferability via Feature Permutation Attacks}

\titlerunning{Transferring Adversarial Attacks Across Heterogeneous Architectures using FPA}

\author{Tao Wu\inst{1}\orcidlink{0000-0002-5927-217X} \and
Tie Luo\inst{2}\thanks{Corresponding author.}\orcidlink{0000-0003-2947-3111}}
\authorrunning{T. Wu and T. Luo}
\institute{Missouri University of Science and Technology, Rolla, MO 65401, USA \and
University of Kentucky, Lexington, KY 40506, USA\\
\email{wuta@umsystem.edu, t.luo@uky.edu}}
\maketitle
\renewcommand\thefootnote{}  
\footnotetext{Advances in Knowledge Discovery and Data Mining. Proceedings of PAKDD 2025.}
\renewcommand\thefootnote{\arabic{footnote}}  

\vspace{-5mm}
\begin{abstract}
Adversarial attacks in black-box settings are highly practical, with transfer-based attacks being the most effective at generating adversarial examples (AEs) that transfer from surrogate models to unseen target models. However, their performance significantly degrades when transferring across {\em heterogeneous} architectures---such as CNNs, MLPs, and Vision Transformers (ViTs)---due to fundamental architectural differences. To address this, we propose {\em Feature Permutation Attack} (FPA), a zero-FLOP, parameter-free method that enhances adversarial transferability across diverse architectures. FPA introduces a novel feature permutation (FP) operation, which rearranges pixel values in selected feature maps to simulate long-range dependencies, effectively making CNNs behave more like ViTs and MLPs. This enhances feature diversity and improves transferability both across heterogeneous architectures and within homogeneous CNNs.
Extensive evaluations on 14 state-of-the-art architectures show that FPA achieves maximum absolute gains in attack success rates of 7.68\% on CNNs, 14.57\% on ViTs, and 14.48\% on MLPs, outperforming existing black-box attacks. Additionally, FPA is highly generalizable and can seamlessly integrate with other transfer-based attacks to further boost {\em their} performance. Our findings establish FPA as a robust, efficient, and computationally lightweight strategy for enhancing adversarial transferability across heterogeneous architectures.

\end{abstract}

\section{Introduction}
Adversarial examples (AEs) pose significant challenges to machine learning security. Transfer-based adversarial attacks \cite{dong2018boosting} are particularly impactful, as they enable black-box attacks \cite{wu23black} by transferring  AEs from white-box surrogate models to unseen target models. However, prior research has focused on transferring within {\em homogeneous} architectures, such as differeent convolutional neural networks (CNNs), where structural similarities between surrogate and target models facilitate effective transfers.

\begin{wrapfigure}{R}{0pt}
\includegraphics[width=0.6\textwidth,trim=2mm 3mm 2mm 1.1cm,clip]{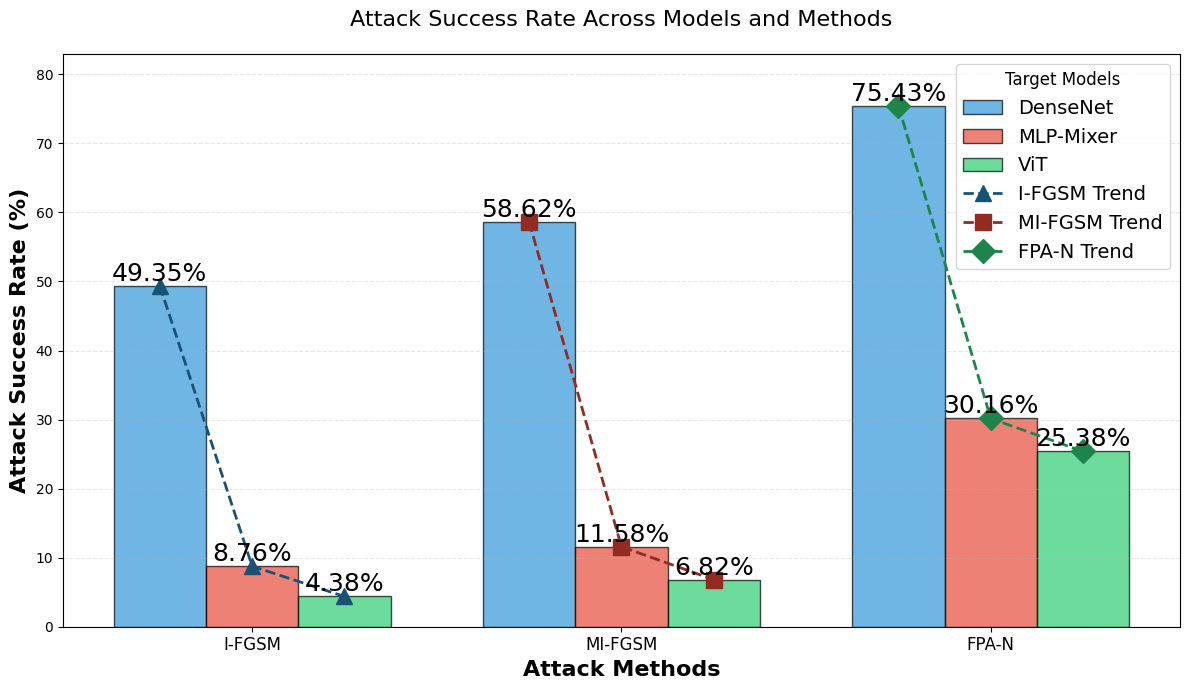}
\caption{Adversarial transferability collapses across heterogeneous architectures. AEs generated on ResNet-50 transfers well to CNNs like DenseNet121 but fail drastically on ViTs and MLPs (tested with I-FGSM and MI-FGSM). Our proposed FPA-N significantly improves this by increasing success rates by $\sim$20 percentage points.}
\label{fig:gap}
\end{wrapfigure}

In constrast, transferring adversarial attacks across heterogeneous architectures (e.g., CNNs, ViTs, MLPs) remains largely ineffective, despite being highly desirable given the growing adoption of non-CNN architectures in computer vision. Indeed, we observe that the adversarial transferability collapses when shifting between heterogeneous models. For instance, as shown in \fref{fig:gap}, AEs generated on ResNet-50 maintain a high success rate (58.62\%) when  transferred to DenseNet121, but the success rate plummets to 6.82\% on ViT-Base~\cite{dosovitskiy2020image} and 11.58\% on MLP Mixer-Base~\cite{tolstikhin2021mlp}, highlighting a critical gap in current transfer-based attack methods. Thus, we pose the following research question: {\it Is it possible to transfer black-box adversarial attacks across heterogeneous architectures effectively?}

To answer this question, we first analyze the fundamental differences between CNNs, ViTs, and MLPs. CNNs extract {\em local} spatial patterns through convolution, leading to relatively {\em small receptive fields}, especially in early layers. ViTs leverage self-attention to model relationships between all token pairs, capturing {\em global dependencies}. MLPs employ fully connected layers to enable comprehensive information exchange across neurons of adjacent layers, effectively modeling {\em long-range connections}. Thus, we hypothesize that poor adversarial transferability stems from CNNs' inadequacy in attending to larger contexts and long-range dependencies, making their generated AEs ineffective against architectures that rely on global feature interactions.

To bridge this gap, we propose {\em Feature Permutation Attack} (FPA), a parameter-free, zero-FLOP method that enhances adversarial transferability by reshuffling feature representations within CNNs. Specifically, we design a feature permutation operation that strategically rearranges pixel values in selected feature maps inside surrogate CNNs. This in effect establishes long-range interactions among input features, making CNNs behave more like ViTs and MLPs. As a result, AEs would be more ``smoothly'' transferred across diverse architectures.

Moreover, FPA is designed to be easily integrated with existing transfer-based attacks without modifying their structures, yet can improve their performance significantly. Our contributions are as follows:
\begin{itemize}
\item We identify the fundamental reasons behind the failure of heterogeneous adversarial transferability, and propose Feature Permutation Attack to address this challenge.
\item FPA is a simple and efficient method requiring zero additional parameters or FLOPs, making it highly efficient.
\item Extensive evaluations on ImageNet using 14 neural architecture and 8 attack methods demonstrate that FPA significantly improves attack success rates (by 8-26 percentage points on average) in the heterogeneous setting, and even improves homogeneous CNN-to-CNN transferability by 2-30 points.
\item FPA is highly generalizable, acting as a ``plug-and-play'' enhancement that integrates with existing transfer-based attacks without requiring modifications to their core structure.
\end{itemize}

\vspace{-2mm}
\section{Related Work}\vspace{-2mm}
\subsection{Transfer-based Attacks on CNNs}\vspace{-2mm}
AEs and their transferability were first identified in CNNs \cite{szegedy2014intriguing}. Since then, various strategies have been developed to enhance their transferability, which can be categorized into the following. (i) Optimization-based methods focus on identifying gradient directions that optimize transferability. For instance, the Momentum Iterative Method (MIM) \cite{dong2018boosting} incorporates a momentum term for stability, while \cite{qin2022boosting,wu2023gnp} targets AEs in regions with uniformly low loss. (ii) Attention-based approaches modify critical features within attention maps, leveraging similarities across different networks. The Attention-guided Transfer Attack (ATA) \cite{wu2020boosting} derives attention maps from objective function gradients, seeking AEs that maximize differences between their attention maps and those of benign samples. Related methods include the Attack on Attention (AoA) \cite{chen2020universal} and Activation Attack (AA) \cite{inkawhich2019feature}. (iii) Input transformation methods smooth gradients by averaging gradients from multiple nearby data points. Notable examples are the Diverse Inputs Method (DIM) \cite{xie2019improving} and Translation-invariant Attack (TIM) \cite{dong2019evading}. (iv) Ensemble methods utilize ensembles of surrogate models, operating under the assumption that if an AE can attack multiple models, it will likely transfer to others as well. Techniques like \cite{gubri2022lgv,wu2024lrs} improve transferability through fine-tuning of pretrained surrogate models.

\vspace{-2mm}
\subsection{Transfer-based Attacks on ViTs}\vspace{-2mm}
Compared to CNNs, transferring AEs between Vision Transformers (ViTs) has been less explored. Recent work \cite{naseer2021improving} dissects a ViT into an ensemble of networks to find multiple discriminative pathways, utilizing class-specific information at each block. \cite{wei2022towards} introduces a dual attack framework, PNA and PatchOut, tailored for ViTs that significantly enhances transferability. \cite{zhou2022improving} presents the Self-Attention Patches Restructure (SAPR), which restructures embedded patches to create diverse connections and maintain regions of interest. The Token Gradient Regularization (TGR) method proposed by \cite{zhang2023transferable} reduces gradient variance in internal ViT blocks. \cite{wang2022generating} introduces the Architecture-oriented Transferable Attacking (ATA) framework, which generates transferable AEs by activating uncertain attention and perturbing sensitive embeddings. Additionally, the Momentum Integrated Gradients (MIG) attack \cite{ma2023transferable} demonstrates improved transferability across different architectures, outperforming existing methods.


\section{Method}\vspace{-2mm}

Given a classifier $f(x): x \in \mathcal{X} \rightarrow y \in \mathcal{Y}$ that outputs a predicted label $y$ for an input $x$, we aim to craft an adversarial example $x^*$ that is visually indistinguishable from $x$ but will be misclassified by the classifier, i.e., $f\left(x^*\right) \neq y$. This objective can be formulated as an optimization problem:
\begin{equation} \label{eq:obj}
    \arg \max _{x^*} \ell\left(x^*, y\right), \;\; \text {s.t. }\left\|x^*-x\right\|_p \leq \epsilon,
\end{equation}
where the loss function $\ell(\cdot, \cdot)$ is often the cross-entropy loss, and the $l_p$-norm measures the discrepancy between $x$ and $x^*$. We consider $p=\infty$ in this work as is common in the literature. I-FGSM is a widely used approach to solve Eq. \ref{eq:obj} by adding a small perturbation iteratively with step size $\alpha$ in the gradient direction:
\begin{equation}
x_{t+1}^{*}=x_t^{*}+\alpha \cdot \operatorname{sign}\left(\nabla_{x_t^{*}} \ell\left(x_t^{*}, y ; \theta\right)\right).
\end{equation}
However, in the black-box setting, the gradient of a loss function w.r.t input $x$ is not accessible. Therefore, the common approach is to leverage an white-box surrogate model to generate AEs that can transfer effectively, i.e., successfully attack as many unseen black-box target models as possible.

\vspace{-2mm}
\subsection{Feature Permutation Attack (FPA)}
As we hypothesize that the inherent limitations of CNNs in capturing long-range dependencies are a primary barrier to heterogeneous adversarial transferability, our proposed Feature Permutation Attack (FPA) introduces long-range feature dependencies via a virtual layer with zero FLOP and zero parameters. This virtual layer can be inserted in any surrogate CNN model before or after any convolutional layer. To formally define the feature permutation (FP) operation, we first express the convolution operation as:
\begin{equation}
Y(m, n, t)=\sum_k \sum_i \sum_j X(m+i, n+j, k) \cdot F(i, j, k, t)
\end{equation}
where $X(m, n, k)$ is the input feature map at spatial location $(m,n)$ for input channel $k$, $Y(m, n, t)$ is the output feature map at spatial location $(m,n)$ for output channel $t$, and $F(\cdot)$ is the filter.
For FPA, we insert a FP layer before layer $l$ in a surrogate model, which reorganizes pixel positions in feature maps with a probability $p$ and a channel ratio $\gamma$. The FP operation $\mathbf{P}(\cdot)$ can be represented as: 
\begin{align}\label{eq:full}
Y(m, n, t)=\sum_k \sum_i \sum_j \mathbf{P}(X(m+i, n+j, k)) \cdot F(i, j, k, t)
\end{align} where
\begin{align}\label{eq:permute}
\mathbf{P}(X(m+i, n+j, k)) = \begin{cases}
\pi X(m+i, n+j, k), & \text{if } k \leq \gamma C \\
X(m+i, n+j, k), & \text{otherwise.}
\end{cases}
\end{align}
Here, $\pi$ is a permutation matrix determined by our permutation strategy (discussed shortly), $C$ is the total number of channels, and $\gamma \in[0,1]$ is a hyper-parameter that controls the proportion of channels to permute. Notably, this permutation operation relies solely on memory operations, which eliminates the need for parameters or floating-point operations (FLOPs), allowing for efficient data rearrangement without the computational overhead as of traditional techniques. Furthermore, our FP layer is differentiable, which allows for smooth integration into PyTorch’s automatic differentiation framework for gradient computation. Therefore, FPA can function as a flexible plug-and-play component for any CNN-based surrogate model.

The permutation strategy $\pi$ is a critical aspect of our method. We propose two permutation strategies in this paper: FPA-R and FPA-N, described below.

\begin{figure*}[t]
    \centering
    \subfloat[{\em Random Permutation:} Pixels within the feature map are rearranged randomly, facilitating global feature interaction and enhancing long-range dependencies.]{
        \includegraphics[width=0.45\textwidth]{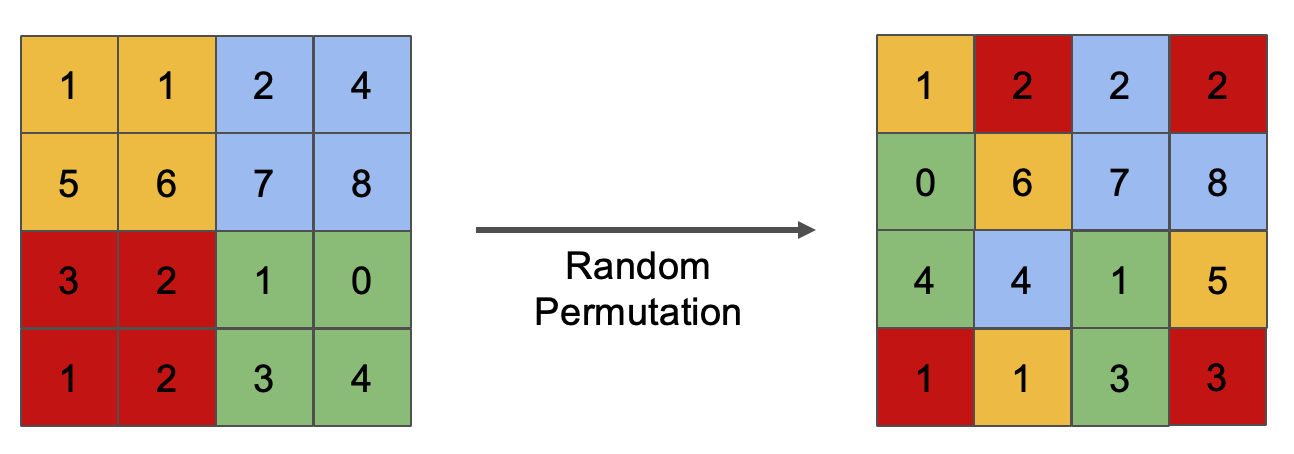}
        \label{fig:image1}
    }
    \hfill
    \subfloat[{\em Neighborhood Permutation:} Each pixel is exchanged with one of its four neighboring pixels, preserving the local spatial relationships and contextual integrity.]{
        \includegraphics[width=0.45\textwidth]{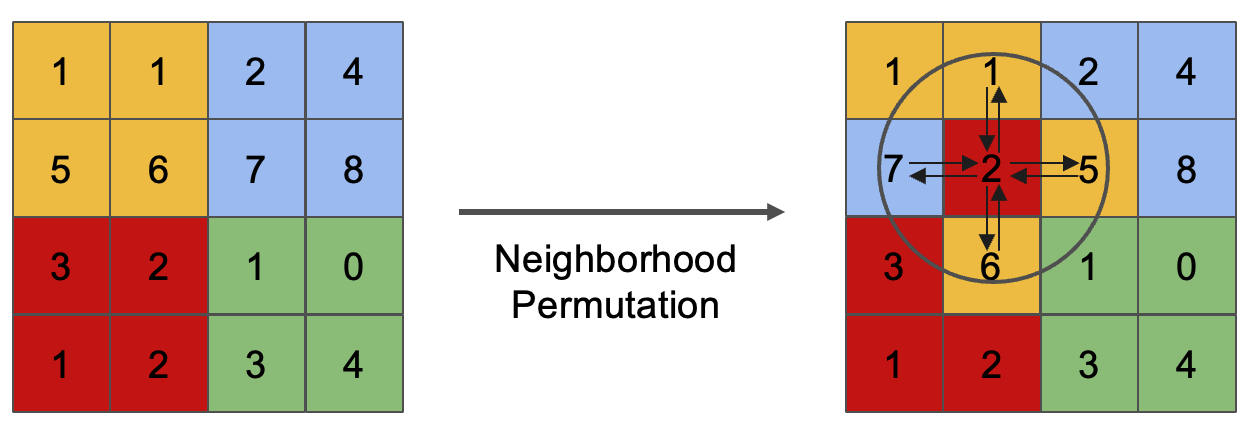}
        \label{fig:image2}
    }
    \caption{Illustration of the two proposed FP strategies: (a) FPA-R enables feature interactions in broader ranges, while (b) FPA-N maintains spatial coherence through local exchanges.}
    \label{fig:FP}\vspace{-5mm}
\end{figure*}

{\bf Random Permutation.}
FPA-R employs random permutation, where pixel positions within each feature map are shuffled randomly during each iteration. This approach treats each pixel independently, introducing a high degree of randomness and unpredictability to the permutation process.

{\bf Neighborhood Permutation.}
FPA-N utilizes neighborhood permutation, where each pixel exchanges its value with one of its four randomly chosen neighboring pixels. This strategy accounts for the spatial relationships of local pixels, resulting in fewer variations compared to FPA-R while preserving the global context of the feature maps.

\begin{wrapfigure}{R}{0pt}
\centering
\includegraphics[width=0.51\textwidth]{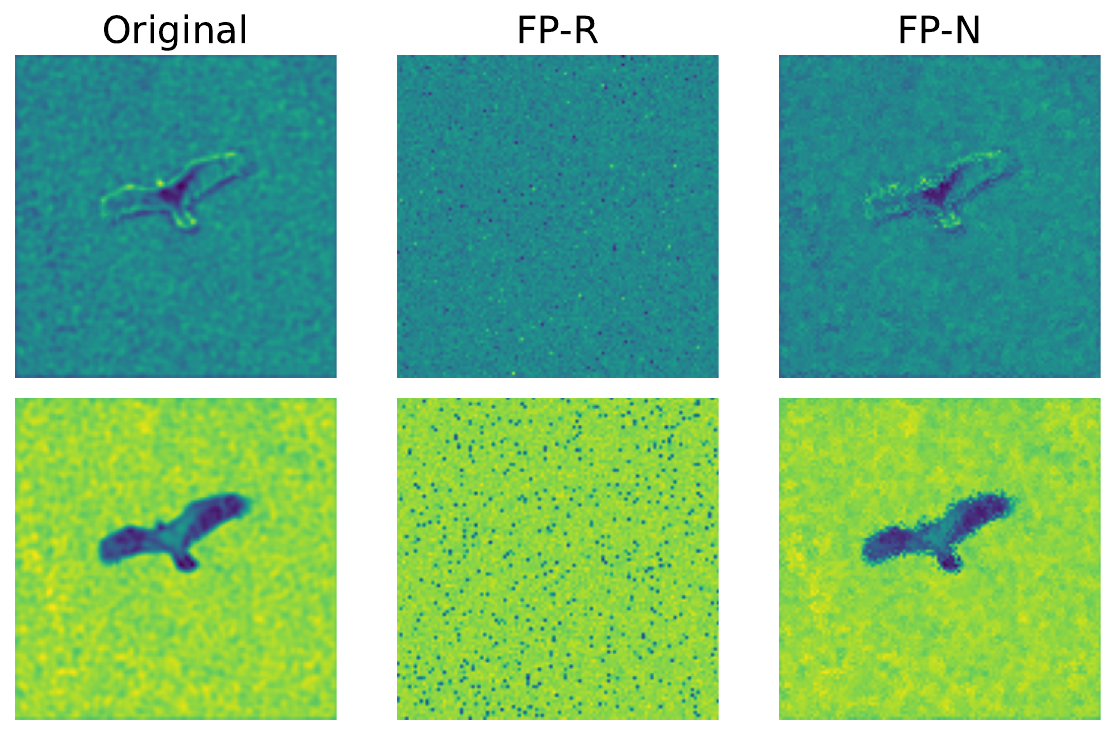}
\caption{Feature maps after applying the two FP strategies with probability $p$.}
\label{fig:feature}
\end{wrapfigure}
\fref{fig:FP} depicts how FPA-R and FPA-N operate on feature maps, and \fref{fig:feature} illustrates the distinct visual patterns resulting from these two strategies applied before the second convolution layer of ResNet-50. On the other hand, it is important to note that both FPA-R and FPA-N facilitate the establishment of long-range feature connections, enabling the CNN surrogate model to behave more like ViTs and MLPs. This alignment enhances adversarial transferability, addressing the challenge posed by architectural differences.

\subsection{Analysis of Why FPA Works}

While our study primarily employs CNN-based surrogate models, FPA is {\em architecture-agnostic} and can be seamlessly applied to other architectures as well. 
In this section, we analyze and explain why FPA would achieve the following desired properties, which will then be evidenced in our experiments.

\textbf{1) Why Better Transferability to Other CNN-Based Target Models?}

We attribute FPA's enhanced effectiveness in transferring AEs across different CNN-based models to three reasons:

{\bf (a) Expanded Receptive Fields:}
FPA enables neurons in early layers to access information beyond their original receptive fields, allowing AEs generated via FPA to leverage global features. While different CNN models may extract slightly varying local features (e.g., texture, color), they often rely on similar global features (e.g., shape) for classification and other tasks. Thus, FPA enhances the transferability of adversarial perturbations.

{\bf (b) Feature Space Transformation:}
Unlike prior methods that focus on input transformations (e.g., resizing, scaling) \cite{xie2019improving,dong2019evading}, FPA adopts transformations directly in the feature space, similar to \cite{li2021simple,bertoin2022local}. This paradigm shift offers several advantages. For comparison, see SIM \cite{lin2020nesterov}, which enhances transferability by averaging gradients across $m$ scaled copies of the input:
\[ \bar{g}_{t+1}=\frac{1}{m} \sum_{i=0}^{m-1} \nabla_{x_t^{*}}\left(\ell\left(x_t^{*} / 2^i, y ; \theta\right)\right). \]
This input transformation increases computation by a factor of $m$. In contrast, FPA operates on the compressed or downsampled feature maps, allowing for more sophisticated variations without additional computation overhead.

{\bf (c) Implicit Ensemble Effect:}
FPA mimics the principle of model ensembling by dynamically altering the (instantiated) convolutional kernels at each iteration through feature permutation. This transforms the surrogate model into a versatile AE generator, exposing the AEs to a broader spectrum of models. Consequently, this reduces overfitting to specific patterns in the surrogate model and promotes better transferability across diverse target models.

\textbf{2) Why Better Transferability to ViT and MLP-Based Target Models?}

FPA enhances adversarial transferability to ViT and MLP-based models by introducing long-range feature dependencies, a core characteristic of these architectures. Through feature permutation, FPA modifies the internal feature representations of CNN-based surrogate models, aligning them more closely with the global context and interaction patterns of ViT and MLP-based models.

More specifically, the creation of long-range connections enables features across different layers to interact more effectively, fostering deeper alignment between the surrogate and target models. This alignment bridges the architectural gap, allowing AEs crafted on CNNs to transfer more successfully to ViTs and MLPs, which inherently rely on global dependencies for classification and other tasks.

\section{Experiments} \label{sec:experiments}\vspace{-2mm}

\subsection{Experimental Setup} 
{\bf Dataset.} We conduct untargeted black-box attacks on classifiers trained on ImageNet, following practices in adversarial attack research \cite{guo2020backpropagating}. A subset of 5,000 correctly classified test images from the ImageNet validation set is randomly selected for evaluation.

\textbf{Models Under Attack.} To evaluate the effectiveness of FPA, we target a diverse set of architectures, including 7 CNNs, 4 ViTs, and 3 MLPs. ResNet-50 serves as the surrogate model, while the target models include:
\begin{itemize}[label=$\bullet$,topsep=1pt]
    \item {\bf CNNs:} VGG-19 \cite{Simonyan2015}, ResNet-152 \cite{he2016deep}, Inception v3 \cite{Szegedy2016}, DenseNet121 \cite{Huang2017densely}, MobileNet v2 \cite{Sandler2018mobilenetv2}, WRN \cite{Zagoruyko2016}, PNASNet \cite{Liu2018}.
    \item {\bf ViTs:} ViT-B \cite{dosovitskiy2020image}, DeiT-B \cite{touvron2021training}, Swin-B \cite{liu2021swin}, BEiT-B \cite{bao2021beit}.
    \item {\bf MLPs:} Mixer-B \cite{tolstikhin2021mlp}, Res-MLP \cite{touvron2022resmlp}, gMLP \cite{liu2021pay}.
\end{itemize}
We follow each model's official pre-processing pipelines for evaluation.

\textbf{Implementation details.} All experiments are conducted using untargeted $\ell_{\infty}$ norm attacks in a black-box setting, with I-FGSM as the base algorithm. Each attack runs for 50 iterations with a perturbation size of $\epsilon=8/255$ and a step size of 2/255, clipping AEs to the valid range [0, 1]. FPA parameters are as follows:
\begin{itemize}[label=$\bullet$,topsep=1pt]
\item {\bf FPA-R:} $\gamma=0.3$, $p=0.2$ (6\% of channels permuted on average), FP layer inserted after the fifth convolutional block.
\item {\bf FPA-N:} $\gamma=0.6$, $p=0.5$, FP layer inserted after the second convolutional block.
\end{itemize}
Results are averaged over 5 independent runs, with all experiments performed on an NVIDIA V100 GPU.

\vspace{-2mm}
\subsection{Experimental Results} 


\begin{table*}[t!]
\begin{center}
\resizebox{0.95\linewidth}{!}{
\renewcommand{\arraystretch}{1}
\begin{tabular}{cccccccc}
\toprule
Method  & VGG-19  & ResNet-152 & Inception-V3 & DenseNet121 & MobileNet-V2 & WRN & PNASNet\\ 
\midrule
I-FGSM  & 43.26\% & 23.65\%    & 21.54\%      & 49.35\%  & 38.21\%   &45.32\% & 18.91\%\\
MI-FGSM  & 52.89\% & 31.56\%    & 32.16\%      & 58.62\%  & 50.35\%   &54.69\%&  29.32\%\\
DIM     & 67.85\% & 41.25\%     & 38.95\%       & 70.26\% & 65.26\%   &68.42\%&  35.46\%\\
TIM     & 46.78\% & 29.14\%      & 27.83\%       & 51.35\%  & 48.31\%  &49.63\%  &  25.34\%      \\
SIM     & 52.82\% & 35.68\%      & 33.68\%       & 58.96\%   & 54.16\%  &58.47\%  &  29.65\%      \\
Admix   & 66.95\% & 43.62\%      & 39.46\%       & 68.47\%  & 59.21\%  &65.61\%&  30.49\%\\
SGM     & 63.46\% & 46.52\%     & 39.26\%      & 71.26\%  & 57.26\%   &64.18\%&  31.25\%\\
LinBP  & 66.31\% & 50.18\%     & 37.89\%      & 69.43\%  & 63.48\%   &68.14\%&  32.06\%\\
\midrule
FPA-R (ours)    & 56.83\% & 43.04\% & 35.62\% & 66.59\% & 58.72\%  &60.84\%&  28.89\%\\
FPA-N (ours)    & \textbf{70.25\%} & \textbf{52.38\%} & \textbf{42.85\%} & \textbf{75.43\%} & \textbf{69.48\%}  & \textbf{72.34\%}&  \textbf{39.74\%}\\
\bottomrule 
\\
\toprule
ViT-B & DeiT-B & Swin-B & BEiT-B & Mixer-B & Res-MLP & gMLP & \textbf{Average} \\\midrule
4.38\% & 4.03\% & 4.96\% & 3.78\% & 8.76\% & 7.94\% & 7.12\%  & 18.99\%\\
6.82\% & 5.86\% & 7.88\% & 6.76\% & 11.58\% & 10.92\% & 11.26\% & 27.83\%\\
10.49\% & 10.35\% & 11.06\% & 12.10\% & 15.68\% & 15.34\%  & 14.82\% & 36.94\%\\
5.23\%  & 5.65\%  & 6.04\% & 4.97\%  & 9.68\%   & 10.03\%  & 8.95\%  & 26.08\%     \\
9.35\%  & 10.23\%  & 10.56\% & 11.05\%   & 11.65\%   & 12.14\% & 10.98\% & 31.79\% \\ 
8.79\% & 9.62\% & 10.26\% & 11.67\% & 13.60\% & 13.43\%  & 13.09\% & 34.63\%\\
11.24\%  & 10.42\%  & 10.96\% & 11.53\%  & 14.82\%   & 15.48\%  & 15.67\% & 36.66\% \\
12.06\%  & 10.36\%   & 11.23\% & 10.85\%   & 14.62\%   & 14.85\% & 15.21\%  & 37.53\%\\ 
\midrule
16.39\% & 14.85\% & 15.68\% & 17.32\% & 18.46\% & 19.15\% & 19.52\%  & 37.70\%\\
\textbf{25.38\%}  & \textbf{24.64\%}& \textbf{25.80\%} & \textbf{26.19\%} & \textbf{30.16\%}  & \textbf{31.43\%}  & \textbf{30.82\%}  & \textbf{45.59\%}\\
\bottomrule
\end{tabular}
} 
\caption{Comparing attack success rates of SOTA transfer-based untargeted attacks on ImageNet. ResNet-50 is the surrogate model. I-FGSM is the backend attack method, under the $\ell_\infty$ constraint ($\epsilon=8/255$).}
\label{tab:comapre_imagenet}
\end{center} 
\end{table*}

\textbf{Comparison with State-of-the-Art (SOTA).}
We compare FPA against 8 SOTA baseline attacks across 14 ImageNet classification models in terms of attack success rate (ASR), as summarized in \tref{tab:comapre_imagenet}. The baselines include a gradient stabilization method (MI-FGSM \cite{Dong2018}), four input-transformation-based attacks (DIM \cite{xie2019improving}, TIM \cite{dong2019evading}, SIM \cite{lin2020nesterov}, and Admix \cite{wang2021admix}), and two model-specific attacks (SGM \cite{wu2019skip} and LinBP \cite{guo2020backpropagating}).

The results demonstrate that FPA-N significantly outperforms all other methods in every scenario. On average, FPA-N achieves an ASR of 45.59\%, exceeding the second-best method (LinBP) by +8.06\%. Even FPA-R, which trails FPA-N, surpasses all existing SOTA methods in overall performance (see the last column of \tref{tab:comapre_imagenet}). Notably, for CNN-based target models, FPA-N outperforms LinBP on PNASNet by +7.68\%. Its advantage becomes even more pronounced against ViT and MLP target models, with improvements of +14.57\% on Swin-B and +14.48\% on Res-MLP compared to the second-best method.

\textbf{Integration with Other Transfer-Based Attacks.}
FPA serves as a plug-and-play technique to enhance existing transfer-based attacks without modifying their core mechanisms. To test this, we integrate FPA with MI-FGSM \cite{Dong2018}, DIM \cite{xie2019improving}, and Admix \cite{wang2021admix}. Results in \tref{tab:combine} reveal substantial improvements in adversarial transferability, with ASR increases of up to +20.04\% (averaged across all 14 target models) when FPA-N is added to MI-FGSM. For instance, when attacking DeiT-B, integrating FPA-N with MI-FGSM achieves a remarkable absolute gain of +22.63\%. These results validate FPA-N as a powerful augmenting tool for boosting the performance of existing black-box attack strategies.

\begin{table*}[t!]
\begin{center}
\resizebox{0.95\linewidth}{!}{
\renewcommand{\arraystretch}{1}
\begin{tabular}{cccccccc}
\toprule
Method  & VGG-19  & ResNet-152 & Inception-V3 & DenseNet121 & MobileNet-V2 & WRN & PNASNet\\ 
\midrule
MI-FGSM           & 52.89\% & 31.56\%    & 32.16\%      & 58.62\%  & 50.35\%   &54.69\%&  29.32\%\\
MI-FGSM + FPA-R  & 66.32\% & 49.13\%    & 45.12\%       &71.56\%  & 65.14\%   &69.10\%&  42.95\%\\
MI-FGSM + FPA-N  & \textbf{75.46\%} & \textbf{57.64\%}     & \textbf{38.95\%}       & \textbf{80.05\%} & \textbf{73.94\%}   &\textbf{78.86\%}& \textbf{49.14\%}\\
\midrule
DIM           & 67.85\% & 41.25\%     & 38.95\%       & 70.26\%   & 65.26\%   &68.42\%&  35.46\%\\
DIM + FPA-R   & 75.61\% & 49.12\%     & 46.35\%       & 76.12\%   & 74.31\%  &77.03\%  & 45.61\%\\
DIM + FPA-N  & \textbf{80.05\%} & \textbf{54.10\%}      & \textbf{50.23\%}       & \textbf{79.96\%}  & \textbf{77.56\%}  &\textbf{82.04\%}&  \textbf{49.34\%}\\
\midrule
Admix           & 66.95\% & 43.62\%    & 39.46\%  & 68.47\%  & 59.21\%  &65.61\%&  30.49\%\\
Admix + FPA-R   & 74.35\% & 48.13\%   & 45.19\%   & 75.49\%  & 68.95\%  &76.01\%&  38.49\%\\
Admix + FPA-N   & \textbf{79.64\%} & \textbf{50.09\%} & \textbf{51.29\%} & \textbf{80.13\%} & \textbf{76.95\%}  & \textbf{81.32\%}&  \textbf{44.68\%}\\
\bottomrule 
\\
\toprule
ViT-B & DeiT-B & Swin-B & BEiT-B & Mixer-B & Res-MLP & gMLP & \textbf{Average} \\\midrule
6.82\% & 5.86\% & 7.88\% & 6.76\% & 11.58\% & 10.92\% & 11.26\% & 27.83\%\\
18.26\% & 18.03\% & 17.95\% & 17.52\% & 21.06\% & 22.16\% & 22.53\% & 39.06\%\\
\textbf{27.95\%} & \textbf{28.49\%} & \textbf{28.65\%} & \textbf{29.33\%} & \textbf{34.02\%} & \textbf{34.57\%}  & \textbf{33.13\%} & \textbf{47.87\%}\\
\midrule
10.49\% & 10.35\% & 11.06\% & 12.10\% & 15.68\% & 15.34\%  & 14.82\% & 36.94\%\\
21.30\%  & 19.16\%  & 18.94\% & 23.15\%  & 24.96\%   & 23.84\% & 25.61\% & 42.94\% \\ 
\textbf{29.65\%} & \textbf{31.49\%} & \textbf{33.16\%} & \textbf{32.09\%} & \textbf{36.16\%} & \textbf{36.98\%}  & \textbf{35.88\%} & \textbf{50.62\%}\\
\midrule
8.79\% & 9.62\% & 10.26\% & 11.67\% & 13.60\% & 13.43\%  & 13.09\% & 34.63\%\\
17.53\%  & 19.23\%  & 20.15\% & 22.36\%   & 25.16\%   & 25.01\% & 24.69\%  & 41.48\%\\ 
\textbf{27.32\%}  & \textbf{28.96\%}& \textbf{30.40\%} & \textbf{33.46\%} & \textbf{32.68\%}  & \textbf{32.92\%}  & \textbf{34.05\%}  & \textbf{53.24\%}\\
\bottomrule
\end{tabular}
} 
\caption{Combining FPA with MI-FGSM, DIM, and Admix, evaluated on ImageNet. ResNet-50 is the surrogate model. I-FGSM is the backend attack method, under the $\ell_\infty$ constraint ($\epsilon=8/255$).}
\label{tab:combine}\vspace{-1cm}
\vspace{-5pt}
\end{center} 
\end{table*} 

\vspace{-2mm}
\subsection{Ablation studies}\vspace{-2mm}
We conducted extensive ablation studies to analyze the impact of key hyperparameters in our FPA approach, focusing on their contributions to adversarial transferability. The results are summarized in \fref{fig:ablation}, with AEs generated using a ResNet-50 surrogate model, and attack success rates averaged across 14 target models.

\vspace{-5mm}
\subsubsection{Ratio of Permuted Channels ($\gamma$)}
The parameter $\gamma$ determines the proportion of feature maps subjected to the FP operation in a given layer. A higher $\gamma$ introduces greater diversity but risks disrupting critical feature structures. As shown in \fref{fig:ablation} (left), FPA-N consistently achieves robust performance across all settings, while FPA-R performs well only when a small ratio of channels is permuted. This indicates that random permutation could disrupt feature structures, reducing effectiveness, whereas neighborhood permutation preserves the feature map's integrity.
\vspace{-5mm}
\subsubsection{Probability of Permutation ($p$)}
The parameter $p$ controls the likelihood that a feature map undergoes the FP operation during each iteration. This stochastic process introduces variability into AE generation, enhancing transferability. As observed in \fref{fig:ablation} (center), FPA-N demonstrates consistent performance improvements across scenarios, while FPA-R benefits from permuting a limited number of channels, ensuring that the core feature structure remains intact.
\vspace{-5mm}
\subsubsection{Position of the FP Layer}
\begin{figure*}[t]
  \begin{minipage}{0.33\textwidth}
    \centering
    \label{fig:ablation_ratio}
    \includegraphics[width=\linewidth]{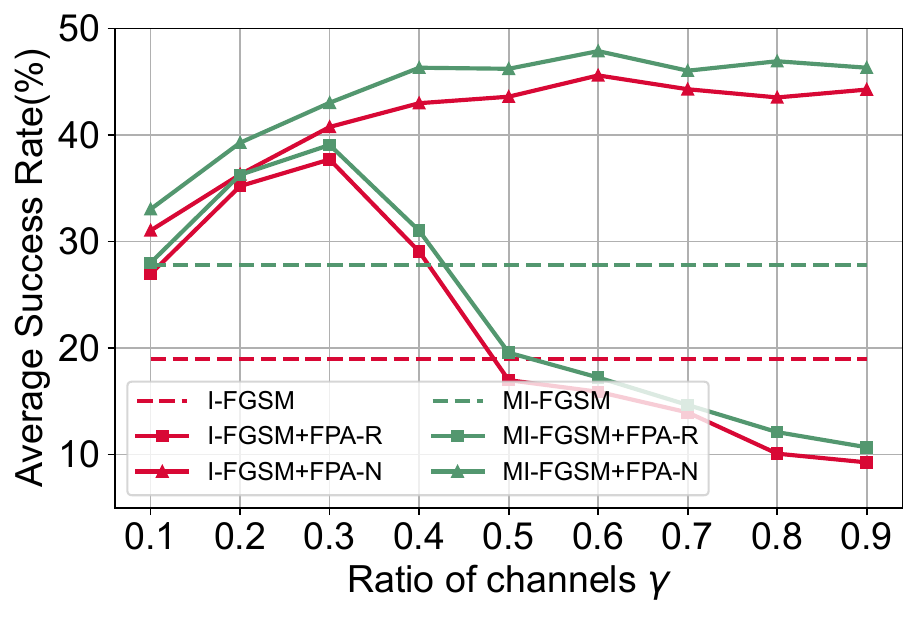}
  \end{minipage}
  \hfill
  \begin{minipage}{0.325\textwidth}
    \centering
    \includegraphics[width=\linewidth]{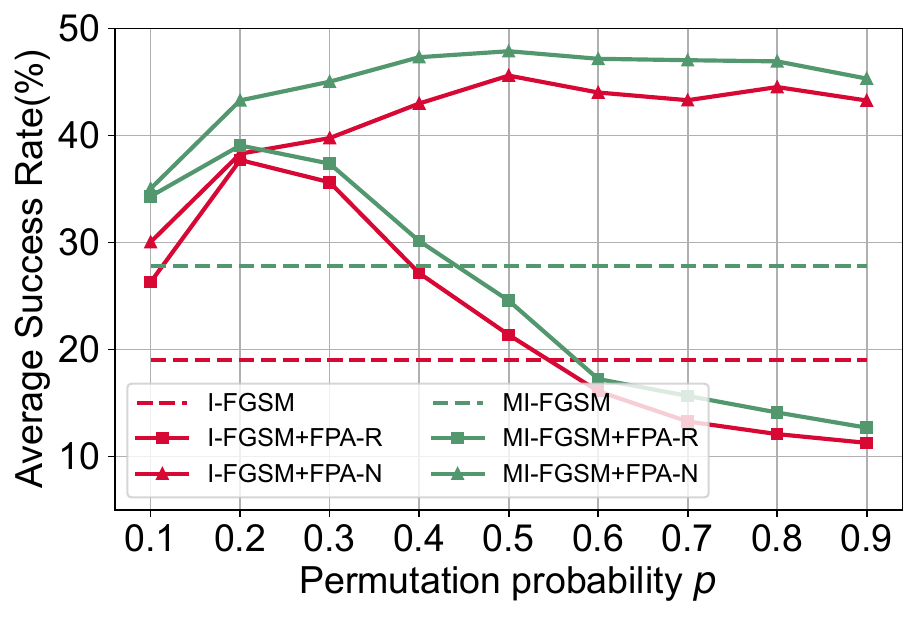}
    \label{fig:ablation_prob}
  \end{minipage}
  \hfill
  \begin{minipage}{0.325\textwidth}
    \centering
    \includegraphics[width=\linewidth]{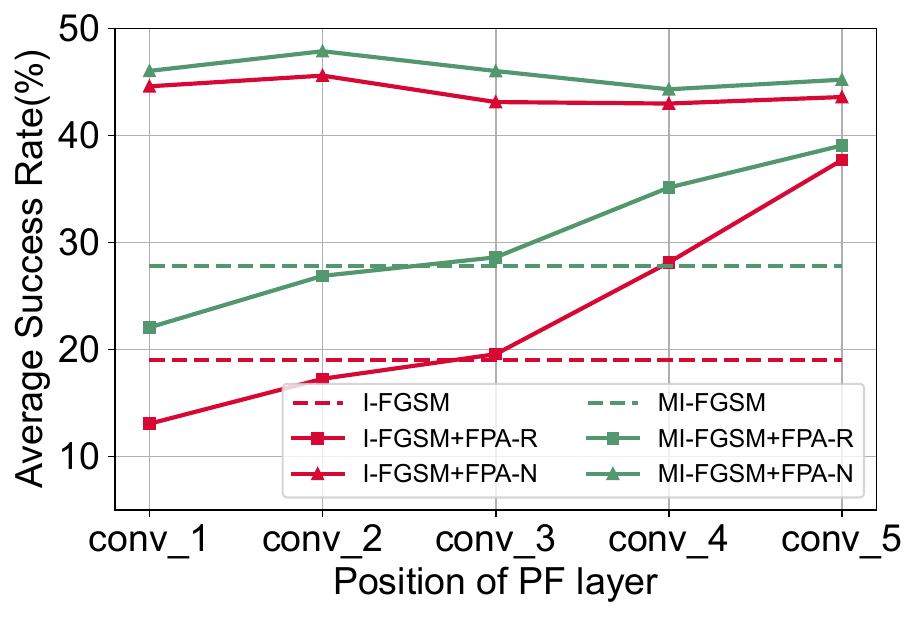}
    \label{fig:ablation_loc}
  \end{minipage}
  \vspace{-3mm}
  \caption{Ablation study on key hyper-parameters of FPA-R and FPA-N: (left) ratio of permuted channels $\gamma$, (center) probability of permutation $p$, and (right) position of the inserted FPA layer $l$.}
  \label{fig:ablation}
\end{figure*}

The location of the FP layer within the network architecture significantly affects performance. \fref{fig:ablation} (right) shows that FPA-R performs better when applied to higher layers, where smaller feature map sizes allow random rearrangement to preserve global feature structures. In contrast, FPA-N achieves stable improvements across different positions due to its ability to maintain local contextual relationships.

\smallskip
\noindent{\bf Key Insights:} Our ablation studies show that FPA-N is not sensitive to hyperparameter variations, such as the ratio of permuted channels $\gamma$ and the probability of permutation $p$, making it robust and easier to tune. The results also demonstrate that neighborhood permutation consistently outperforms random permutation, as the former better captures and preserves local contextual information. These findings underscore the importance of carefully designing feature permutation strategies to optimize adversarial transferability.

\vspace{-2mm}
\subsection{Wall Clock Runtime}\vspace{-2mm}

Finally, we also evaluated the computational efficiency of our proposed permutation operation, which is executed solely through memory operations without requiring matrix computations, additional parameters, or FLOPs. This design makes FPA efficient.

\smallskip
\begin{table}[h]
\centering
\resizebox{0.75\linewidth}{!}{
\begin{tabular}{ccccccccccc}
\toprule
Methods &I-FGSM  & MI-FGSM  & DIM & TIM & SIM & Admix & SGM  & FPA-R   & FPA-N \\\midrule
Time (mins)   & 4.2 & 4.9 & 5.9 & 6.7 & 21.6 & 15.3 & 4.5  & 4.2 & 4.3 \\
\bottomrule
\end{tabular}}\vspace{2pt}
\caption{Comparing wall clock runtime for FPA and baseline attacks on ImageNet.}
\label{tab:time}\vspace{-3mm}
\end{table}

To validate this, we recorded the wall clock runtime, with results presented in \tref{tab:time}. The experiments, conducted on ImageNet (5,000 images) using ResNet-50 as the surrogate model on an NVIDIA RTX 3080 GPU, confirm that our methods introduce minimal overhead. This demonstrates the practicality of FPA in real-world scenarios, offering significant improvements in adversarial transferability without compromising efficiency.

\section{Conclusion}\label{sec:conclusion} \vspace{-2mm}

This paper tackles the critical challenge of adversarial transferability across heterogeneous architectures of surrogate and target models. To address this issue, we propose Feature Permutation Attack (FPA)---a simple, efficient, and highly effective solution. FPA strengthens long-range feature dependencies by strategically permuting pixel values within selected feature maps, all without introducing additional parameters or FLOPs.

FPA improves adversarial transferability in two key ways: (1) by increasing feature diversity, it enhances transferability across CNN-based target models, and (2) by emulating the global feature modeling behavior of ViTs and MLPs, it facilitates better transferability from CNNs to these architectures. Extensive experiments confirm FPA’s effectiveness, demonstrating substantial gains in attack success rates against state-of-the-art deep neural networks and robust defense mechanisms.

With its resource-efficient, architecture-agnostic, and plug-and-play design, FPA stands out as a versatile enhancement to existing transfer-based attack strategies, offering a promising direction for future adversarial research.


\bibliographystyle{splncs04}
\bibliography{ref}

\end{document}